\documentclass[sigconf]{acmart}
\usepackage{caption}
\usepackage{subcaption} %
\usepackage{cleveref}
\usepackage{array}
\usepackage{makecell}
\usepackage{todonotes}
\usepackage{mdframed}
\usepackage{multirow}

\usepackage{pgfplots}
\pgfplotsset{compat=1.18} %
\usetikzlibrary{pgfplots.colormaps} %

\AtBeginDocument{%
  }

\setcopyright{acmlicensed}
\copyrightyear{2018}
\acmYear{2018}
\acmDOI{XXXXXXX.XXXXXXX}
\acmConference[SIGSPATIAL '25]{International Conference on Advances in Geographic Information Systems}{June 03--05, 2025}{Woodstock, NY}
\acmISBN{978-1-4503-XXXX-X/2018/06}

\begin{document}

\title{HydroChronos: Forecasting Decades of Surface Water Change}

\author{Daniele Rege Cambrin}
\orcid{1234-5678-9012}
\authornotemark[1]
\affiliation{%
  \institution{Politecnico di Torino, Italy}
  \city{}
  \country{}
}
\email{daniele.regecambrin@polito.it}

\author{Eleonora Poeta}
\orcid{}
\affiliation{%
  \institution{Politecnico di Torino, Italy}
  \city{}
  \country{}
}
\email{eleonora.poeta@polito.it}

\author{Eliana Pastor}
\affiliation{%
  \institution{Politecnico di Torino, Italy}
  \city{}
  \country{}
}
\email{eliana.pastor@polito.it}

\author{Isaac Corley}
\affiliation{%
  \institution{Wherobots, USA}
  \city{}
  \country{}
  }
\email{isaac@wherobots.com}

\author{Tania Cerquitelli}
\affiliation{%
  \institution{Politecnico di Torino, Italy}
  \city{}
  \country{}
}
\email{tania.cerquitelli@polito.it}

\author{Elena Baralis}
\affiliation{%
  \institution{Politecnico di Torino, Italy}
  \city{}
  \country{}
  }
\email{elena.baralis@polito.it}

\author{Paolo Garza}
\affiliation{%
  \institution{Politecnico di Torino, Italy}
  \city{}
  \country{}
  }
\email{paolo.garza@polito.it}

\renewcommand{\shortauthors}{Rege Cambrin et al.}

\begin{abstract}
 Forecasting surface water dynamics is crucial for water resource management and climate change adaptation. However, the field lacks comprehensive datasets and standardized benchmarks. In this paper, we introduce \textsc{HydroChronos}, a large-scale, multi-modal spatiotemporal dataset for surface water dynamics forecasting designed to address this gap. We couple the dataset with three forecasting tasks. The dataset includes over three decades of aligned Landsat 5 and Sentinel-2 imagery, climate data, and Digital Elevation Models for diverse lakes and rivers across Europe, North America, and South America. We also propose AquaClimaTempo UNet, a novel spatiotemporal architecture with a dedicated climate data branch, as a strong benchmark baseline. Our model significantly outperforms a Persistence baseline for forecasting future water dynamics by +14\% and +11\% F1 across change detection and direction of change classification tasks, and by +0.1 MAE on the magnitude of change regression. Finally, we conduct an Explainable AI analysis to identify the key climate variables and input channels that influence surface water change, providing insights to inform and guide future modeling efforts.
\end{abstract}

\begin{CCSXML}
<ccs2012>
   <concept>
       <concept_id>10010147.10010178.10010224</concept_id>
       <concept_desc>Computing methodologies~Computer vision</concept_desc>
       <concept_significance>500</concept_significance>
       </concept>
   <concept>
       <concept_id>10010405.10010432.10010437.10010438</concept_id>
       <concept_desc>Applied computing~Environmental sciences</concept_desc>
       <concept_significance>500</concept_significance>
       </concept>
   <concept>
       <concept_id>10002951.10003227.10003236.10003237</concept_id>
       <concept_desc>Information systems~Geographic information systems</concept_desc>
       <concept_significance>500</concept_significance>
       </concept>
 </ccs2012>
\end{CCSXML}

\ccsdesc[500]{Computing methodologies~Computer vision}
\ccsdesc[500]{Applied computing~Environmental sciences}
\ccsdesc[500]{Information systems~Geographic information systems}

\keywords{Spatiotemporal Forecasting, Surface Water Dynamics, Remote Sensing, Explainable AI, Multi-modal Data Fusion}
\begin{teaserfigure}
  \centering
    \centering
    \includegraphics[width=0.8\linewidth]{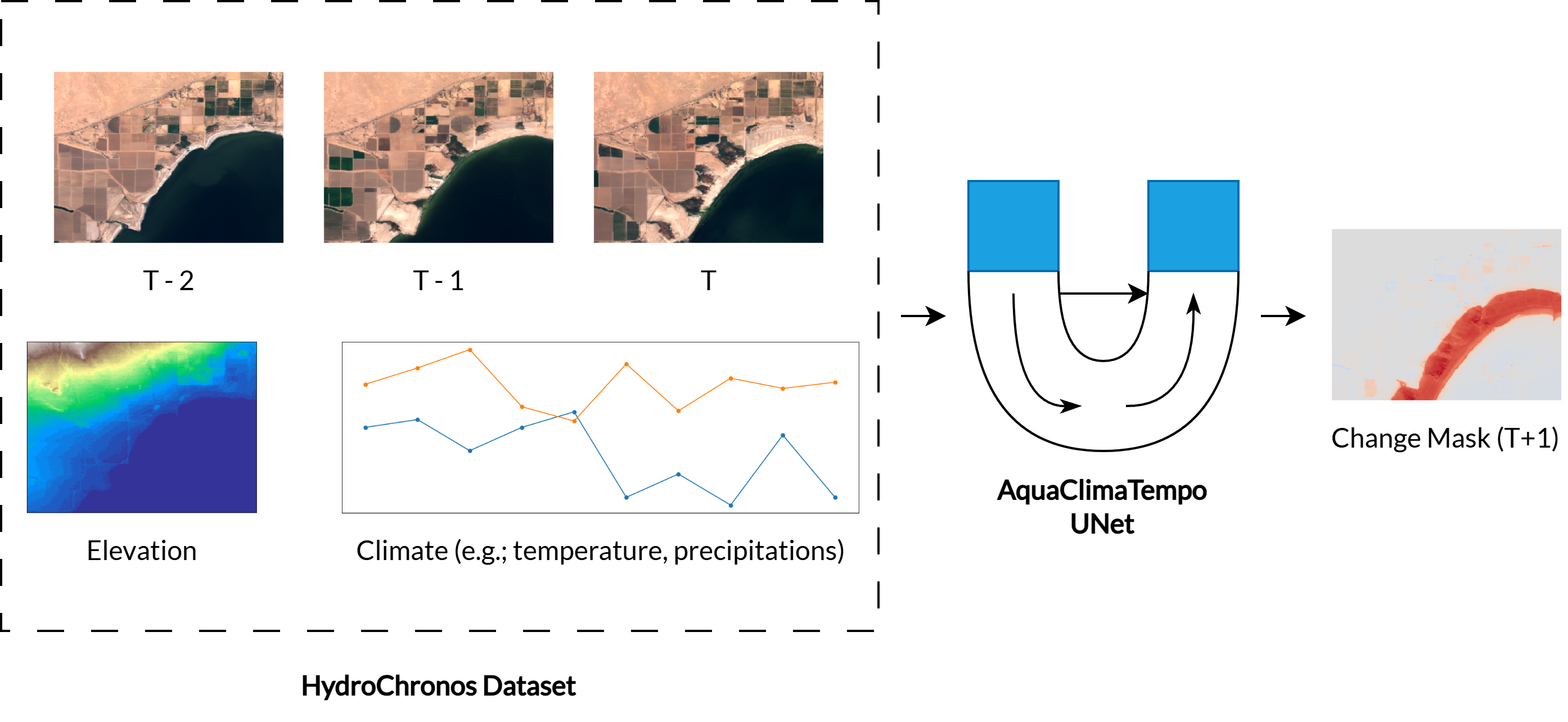}
  \caption{Forecasting future water landscapes: The \textsc{HydroChronos} dataset (left) provides rich multi-modal inputs including satellite image series, elevation, and climate data. Our \textit{AquaClimaTempo UNet} (center) processes this information to predict future surface water dynamics, generating a \textit{change mask} for the forecasted water dynamics for the next timestep (T+1) (right).}
  \label{fig:teaser}
\end{teaserfigure}

\received{6 June 2025}

\maketitle

\section{Introduction}
The escalating global challenges of water scarcity, climate change, and their profound impacts on ecosystems and human societies underscore the critical importance of understanding and forecasting surface water dynamics \cite{Sunkaia2023,Wang2023}. Effective water resource management for agriculture, energy, and consumption relies on predicting future water availability. As climate change intensifies droughts and floods, robust forecasting models are indispensable for adaptation, enabling interventions and improving resilience \cite{Dikshit2021,Jones2023}. Spatio-temporal variability of surface water requires advanced analytics to capture complex hydrological processes and environmental responses \cite{Wang2025}. 
Despite the pressing need for accurate and long-term surface water predictions, the research landscape is hampered by significant limitations. A primary obstacle is the lack of comprehensive, large-scale datasets specifically curated for forecasting tasks. Satellite observations are often fragmented or not integrated with crucial climate and topographic data. Consequently, there is also a scarcity of well-defined predictive tasks that leverage these multi-modal data sources to forecast surface water dynamics.

To fill the gap, this paper introduces \textsc{HydroChronos}, a novel, large-scale, multi-modal spatio-temporal dataset specifically designed to foster research in surface water forecasting. \textsc{HydroChronos} is characterized by an extensive temporal coverage, encompassing over three decades of Landsat 5 and Sentinel-2 satellite imagery. This imagery is integrated with corresponding climate variables (e.g., precipitation and temperature) and a Digital Elevation Model (DEM) for a diverse set of lake and river systems across Europe, the United States, and Brazil. 
Using this dataset, we define three standardized predictive tasks for surface water dynamics forecasting from satellite imagery (optionally with climate data): binary change detection, direction of change classification, and magnitude of change regression.

Building upon the \textsc{HydroChronos} dataset and the defined forecasting tasks, we propose a robust baseline model: AquaClimaTempo UNet (ACTU). This model, based on UNet \cite{Ronneberger2015} and ConvLSTM \cite{shi2015}, features a climate data branch to learn interactions between historical water dynamics and climatic drivers. Our experimental results demonstrate that this model significantly outperforms the commonly used persistence baseline in forecasting future water dynamics. Furthermore, to foster transparency and understanding, we conduct an Explainable AI (XAI) analysis. This analysis offers insights into model decisions, identifies key drivers of surface water changes, and guides future research.

The contributions of this paper (\Cref{fig:teaser}) can be summarized as follows:
\begin{itemize}
    \item We introduce \textsc{HydroChronos}, the first dataset tailored for water dynamics forecasting, including remote-sensed images, climate variables, and DEM.
    \item We introduce three tasks of surface water dynamics forecasting, offering a benchmark for future research in spatio-temporal predictive modeling
    \item We introduce ACTU as a baseline model with the possibility to integrate climate variables and DEM
    \item We performed an XAI analysis on our models to guide future research and understand the influence of various factors
\end{itemize}

The code and the dataset are available for reproducibility at \url{https://github.com/DarthReca/hydro-chronos}.

\section{Related Work}
Forecasting surface water dynamics requires advancements in satellite monitoring, time-series analysis, spatio-temporal modeling, climate data integration, and interpretability. This section reviews existing literature across these domains, contextualizing the contributions of \textsc{HydroChronos} and our proposed methodology.

\subsection{Surface Water Monitoring}
Satellite remote sensing revolutionized monitoring surface water extent and dynamics across vast scales and diverse temporal resolutions. Landsat \cite{Wulder2022} and the Sentinel \cite{drusch2012} missions have provided decades of optical imagery, forming the backbone of many surface water mapping efforts. Common water delineation methods include spectral indices like the Normalized Difference Water Index (NDWI) \cite{mcfeeters1996} and the Modified NDWI (MNDWI) \cite{xu2006}, leveraging water's spectral reflectance. Machine learning classifiers have also been widely employed for more accurate and robust water body extraction \cite{Feyisa2014,Cao2024}. These efforts have culminated in the development of several large-scale and global surface water datasets \cite{Pekel2016,Yamazaki2015}.

These datasets offer insights into past water dynamics but focus on retrospective analysis, not forecasting, providing only the masks of water extents over time. Moreover, they are not explicitly structured for the development and validation of long-term predictive models that integrate auxiliary drivers like climate. Additionally, they are still obtained from automatic extraction, and so their accuracy is dependent on the employed model. \textsc{HydroChronos} is specifically designed for multi-year surface water dynamics forecasting, integrating in a single dataset, imagery, climate variables, and DEM. This multi-modal structure, curated for diverse hydrological systems, provides a rich foundation for developing generalizable models.

\subsection{Time-Series Forecasting in Hydrology}

The analysis and forecasting of hydrological variables like streamflow or water levels was studied for a long time \cite{Machiwal2012}. Both traditional machine learning approaches \cite{Verbesselt2010} and deep learning-based (e.g, based on Long Short-Term Memory \cite{Hochreiter1997}) have been applied in Earth observation data \cite{Reichstein2019,russwurm2017}. While recent advancements in remote sensing posed the problem of forecasting a snapshot image of the future, including exogenous variables \cite{Benson2024}, no application can be seen in hydrology, to the best of our knowledge. Forecasting surface water dynamics is influenced by complex, non-linear interactions between past states, seasonality, and external drivers. \textsc{HydroChronos} proposes to fill the gap, enabling scientists to experiment with a large-scale corpus tailored for hydrological applications based not only on image data, but also on exogenous climate variables.

Deep learning architectures have demonstrated remarkable success in a wide array of environmental modeling and Earth observation tasks, thanks to their ability to learn hierarchical features from large, complex datasets. U-Net architectures \cite{Ronneberger2015} still prove to be a strong baseline for semantic segmentation of satellite imagery \cite{corley2024,Cambrin2023}, including water body delineation \cite{Cao2024}, and more recently, for spatio-temporal forecasting tasks where the output is an image or a sequence of images \cite{Weyn2020}. Our AquaClimaTempo UNet (ACTU), building on similar architectures\cite{Benson2024,shuting2022,Kamangir2024}, adds a dedicated branch for time-series climate data integration and gated fusion to balance climate and optical features. This allows the model to learn how climatic factors modulate surface water dynamics, moving beyond purely auto-regressive image forecasting.

\subsection{Explainable AI in Earth Sciences} %

The complexity of AI models in critical domains like Earth sciences demands transparency and interpretability \cite{Adadi2018,samek2019}. XAI techniques have been applied to environmental models to identify key input features or understand model behavior~\cite{Yang2024}. 
In our case, the integration of climate variables such as precipitation, temperature, and evapotranspiration is well-established as essential for accurate hydrological modeling \cite{Gleick1987,Chiew2002}. Combining climate data with satellite observations presents significant opportunities but also challenges, including issues of scale mismatch, data assimilation, and capturing complex, potentially lagged, interactions. 
Prior studies focus primarily on predictive accuracy, often overlooking interpretability. 
In contrast, our approach incorporates XAI to analyze the relative importance of historical spatio-temporal patterns and climate drivers in predicting future surface water changes.

\section{HydroChronos Dataset}
\begin{figure}[htb] %
    \centering
    \begin{subfigure}[b]{0.49\linewidth} %
        \centering
        \includegraphics[height=4.5cm]{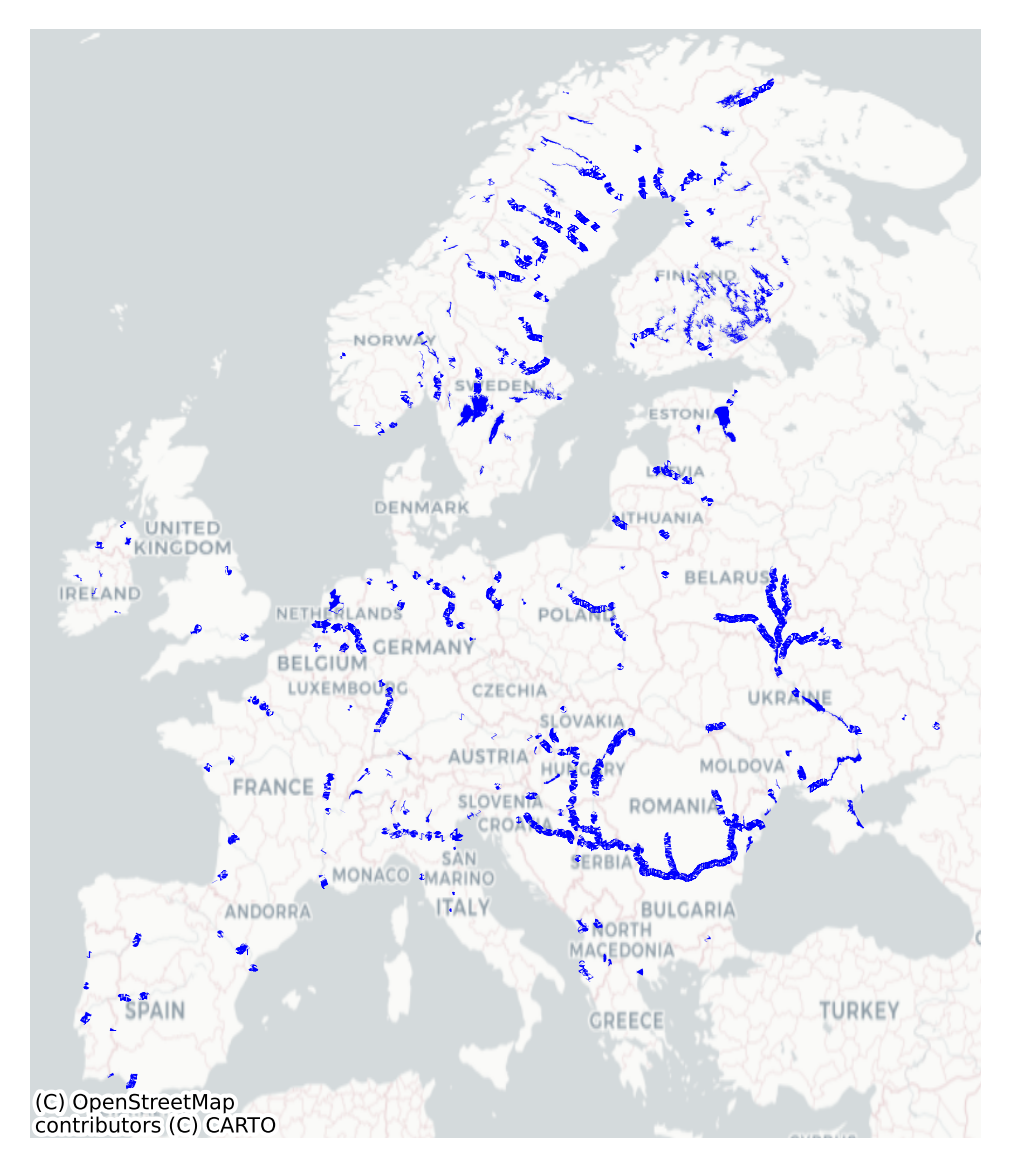} 
        \caption{Europe}
        \label{fig:sub:europe}
    \end{subfigure}
    \hfil %
    \begin{subfigure}[b]{0.49\linewidth}
        \centering
        \includegraphics[height=4.5cm,keepaspectratio]{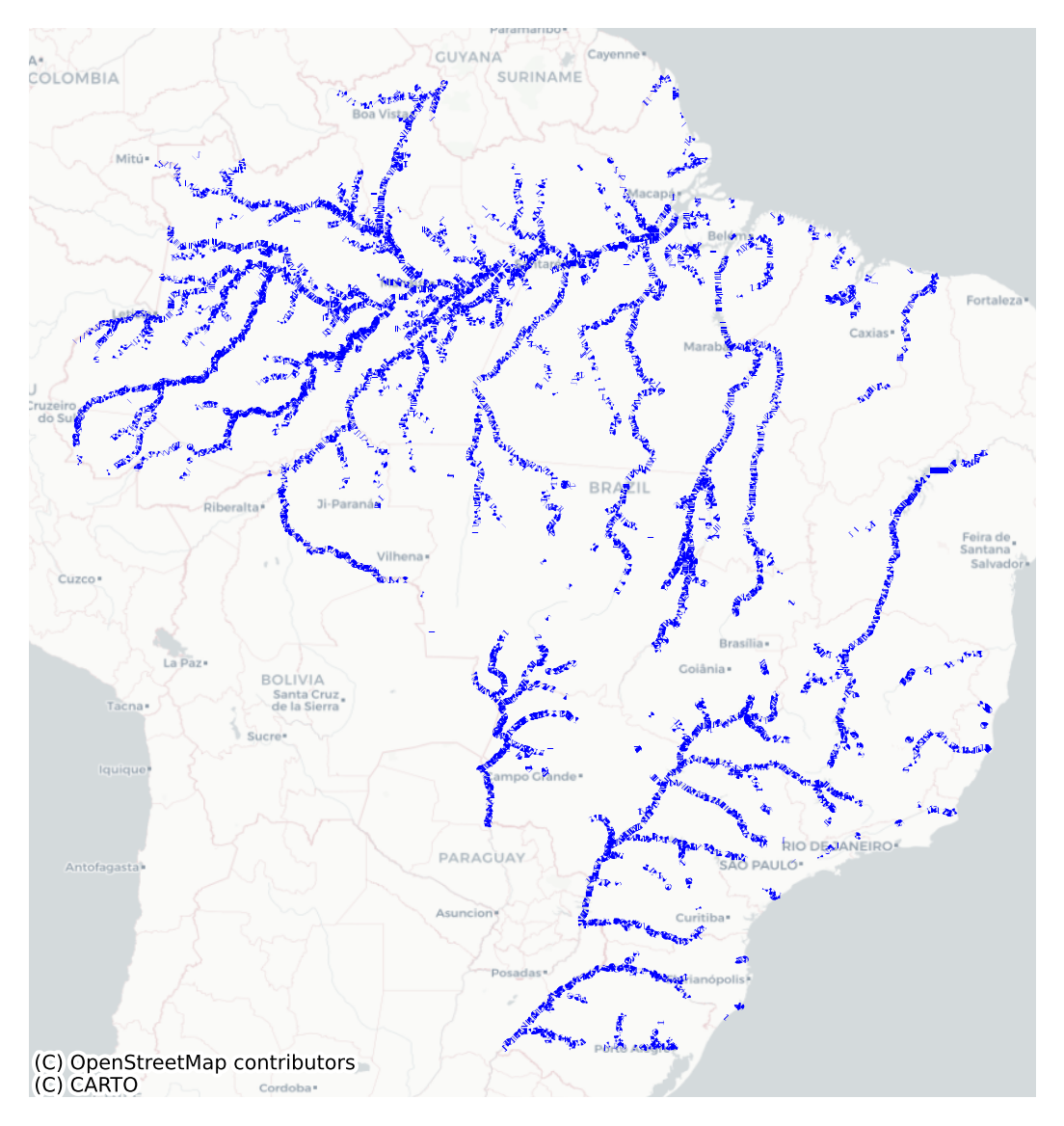}
        \caption{Brazil}
        \label{fig:sub:brazil}
    \end{subfigure}

    \vspace{1ex} %

    \begin{subfigure}[b]{0.9\linewidth} %
        \centering
        \includegraphics[width=\textwidth]{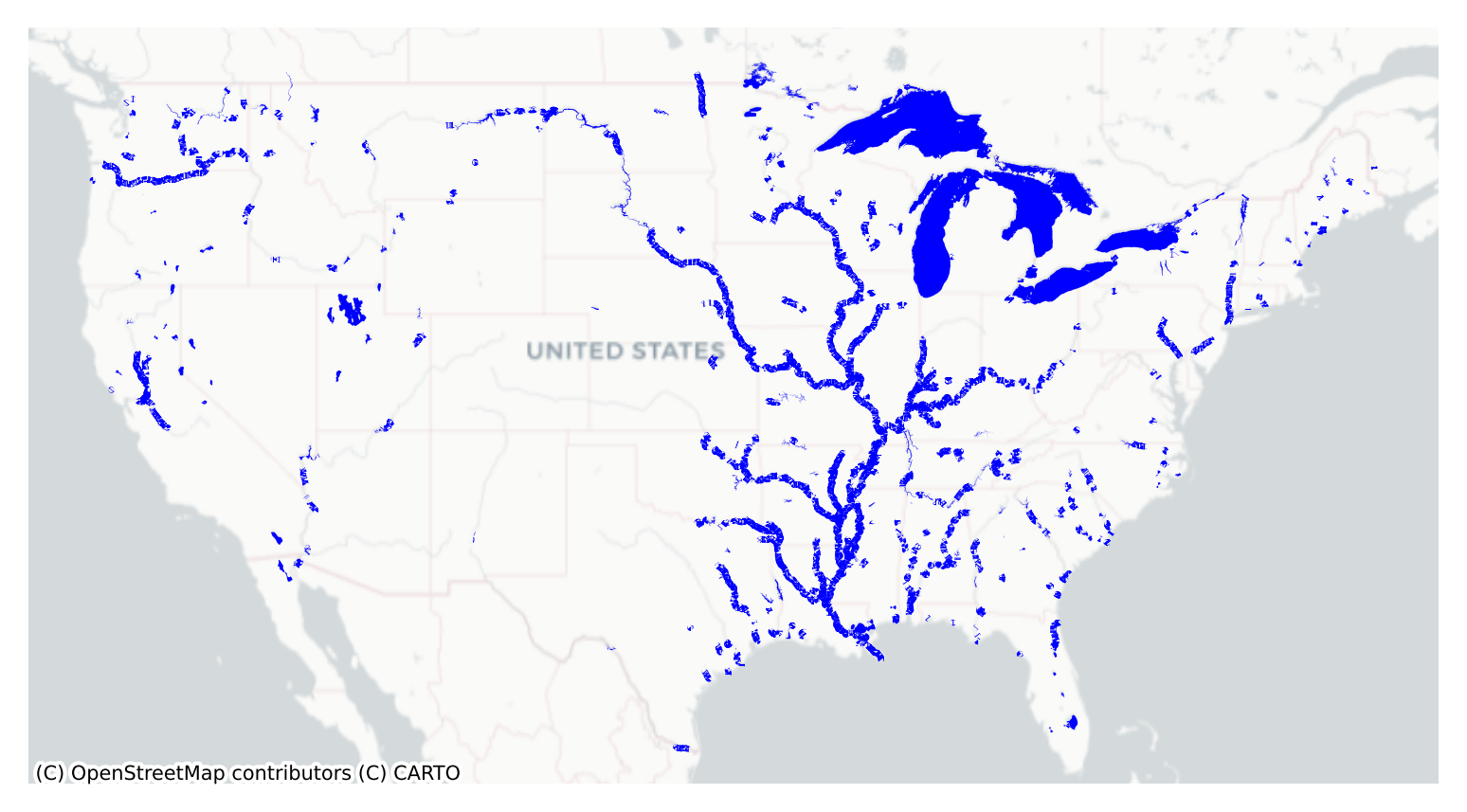}
        \caption{United States}
        \label{fig:sub:unitedstates}
    \end{subfigure}
    \caption{Distribution of lakes and rivers in \textsc{HydroChronos}}
    \label{fig:lakes_rivers_map}
\end{figure}

In this section, we present the newly created dataset: HydroChronos. The dataset is composed of time series of images from Landsat-5 and Sentinel-2, time series of climate variables from TERRACLIMATE \cite{Abatzoglou2018}, and a DEM. The selected lakes' and rivers' names are derived from HydroLAKES \cite{Messager2016} and HydroRIVERS \cite{Lehner2013} and cover USA, Europe, and Brazil as shown in \Cref{fig:lakes_rivers_map}.

\subsection{Landsat-5 and Sentinel-2 images}
To capture long-term changes and recent dynamics, \textsc{HydroChronos} uses imagery from Landsat-5 and Sentinel-2. Sentinel-2 provides imagery with superior spatial resolution (10m/20m) and spectral quality compared to Landsat-5 (30m). However, its temporal coverage is limited to the period from 2015 to 2024. To extend the historical perspective, we include Landsat-5 imagery, which covers the period from 1990 to 2010. 

To ensure data quality, we selected Top-Of-Atmosphere (TOA) images with the lowest cloud coverage possible, prioritizing clear observations of water bodies. To ensure comparable hydrological conditions and minimize unrelated seasonal variability, we selected May-August images (Northern Hemisphere summer).

Recognizing the spectral differences between the two sensors, we selected a consistent set of spectral bands. Sentinel-2 provides up to 13 spectral bands, while Landsat-5 offers 7. We harmonized the data by selecting 6 comparable bands that are available on both sensors, as shown in \Cref{tab:bands}. All imagery is provided at a spatial resolution of 30m and projected to WGS84. An RGB version of a sample can be seen in \Cref{fig:rgb_sample}.

\begin{table}[htb]
    \centering
    \caption{Landsat (L) and Sentinel (S) coupled bands included in the dataset. \textit{NIR} is Near InfraRed and SWIR is \textit{Short-Wave InfraRed}}
    \label{tab:bands}
    \begin{tabular}{cc|cc}
        \toprule
         Landsat & Sentinel & Description & Central Wavelength (L/S) \\\midrule
         B1 & B2 & Blue & 485/492 nm \\
         B2 & B3 & Green & 560/560 nm  \\
         B3 & B4 & Red & 660/665 nm \\
         B4 & B8 & NIR & 830/833 nm \\
         B5 & B11 & SWIR & 1650/1610 nm \\
         B7 & B12 & SWIR & 2220/2190 nm \\\bottomrule
    \end{tabular}
\end{table}

\subsection{Digital Elevation Model}
A static Digital Elevation Model (DEM) provides essential topographic context for hydrological analysis. The DEM for \textsc{HydroChronos} is sourced from the Copernicus GLO30 DEM \cite{copernicus_dem} dataset, which provides global coverage at a spatial resolution of approximately 30 meters. A sample can be seen in \Cref{fig:dem_sample}. This single-timestep layer captures the terrain elevation for each study area, crucial for tasks such as watershed delineation, flow accumulation analysis, and understanding the topographical influence on water body characteristics \cite{Novoa2015}.

\subsection{Climate Variables}
\begin{figure}[htb]
    \centering
    \begin{subfigure}[b]{0.4\linewidth}
        \centering
        \includegraphics[width=\linewidth]{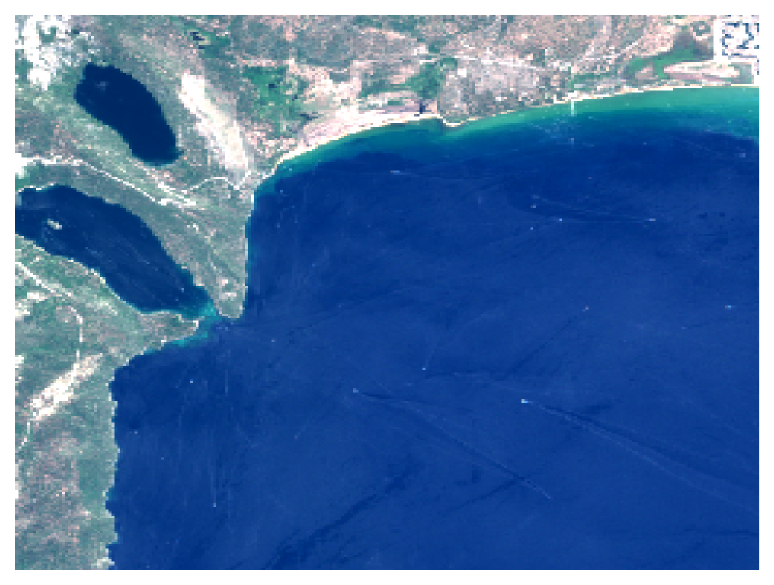}
        \caption{RGB}
        \label{fig:rgb_sample}
    \end{subfigure}
    \begin{subfigure}[b]{0.4\linewidth}
        \centering
        \includegraphics[width=\linewidth]{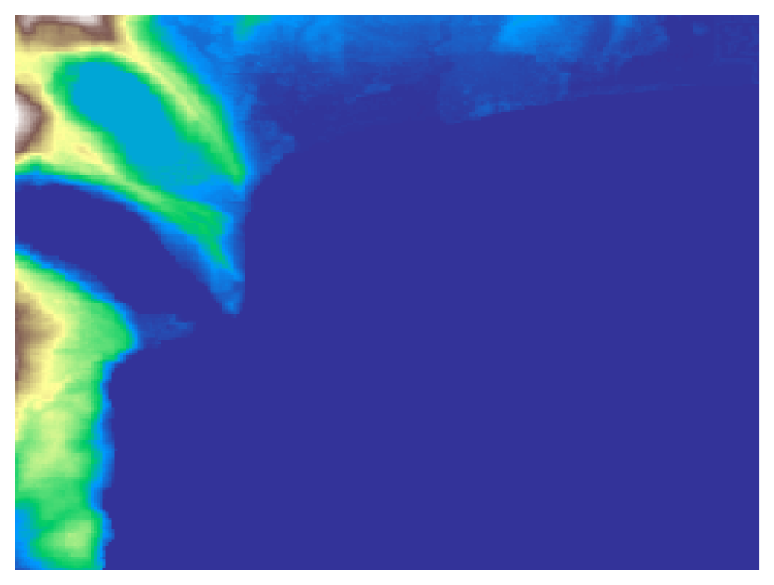}
        \caption{DEM}
        \label{fig:dem_sample}
    \end{subfigure}

    \vspace{5mm}
    
    \begin{subfigure}[b]{0.9\linewidth}
        \centering
        \includegraphics[width=\linewidth]{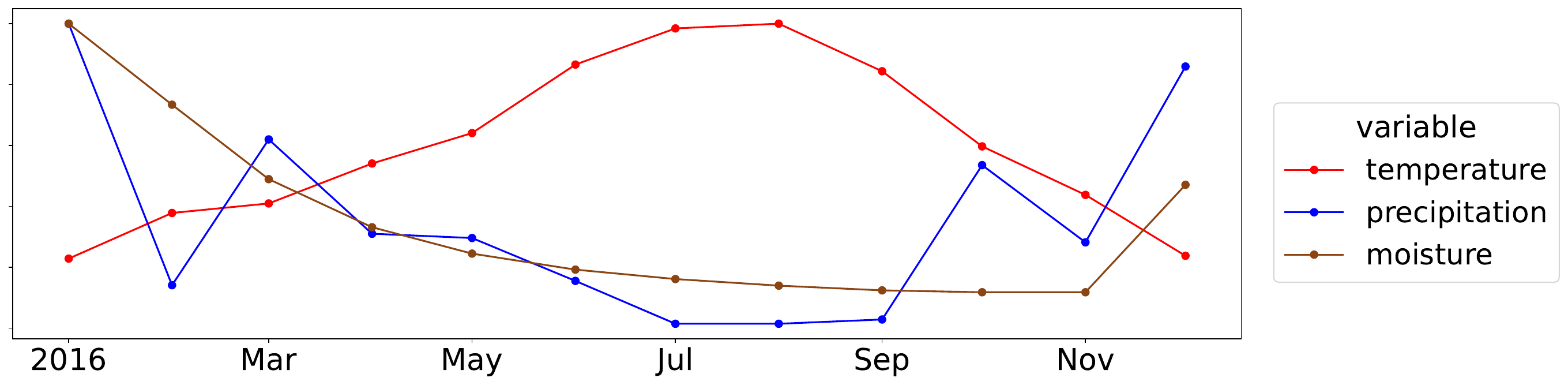}
        \caption{Climate timeseries}
        \label{fig:climate_sample}
    \end{subfigure}
    \caption{Sample in the three modalities: optical (RGB channels only for visualization), DEM, and climate.}
    \label{fig:enter-label}
\end{figure}

To complement the remote sensing data with key environmental drivers, \textsc{HydroChronos} includes time series of climate variables from the TERRACLIMATE \cite{Abatzoglou2018} dataset. It provides monthly climate data globally at a resolution of approximately 4.6 km. The dataset includes 14 variables: actual evapotranspiration, climate water deficit, reference evapotranspiration, precipitation accumulation, runoff, soil moisture, downward surface shortwave radiation, snow water equivalent, maximum temperature, minimum temperature, vapor pressure, vapor pressure deficit, Palmer Drought Severity Index, and wind speed at 10m. We include the complete monthly time series for the periods corresponding to the imagery: 1990-2010 and 2015-2024. A subset of these time series can be seen in \Cref{fig:climate_sample}. These climate variables can be used to analyze the relationship between climatic conditions and observed changes in water bodies captured by the satellite imagery.

\subsection{Splits}
Since each water basin has its peculiar behavior, which could be difficult to summarize in simple hydrological variables even if two of them are near, the most straightforward way is to generalize temporally instead of trying to generalize spatially. We use the Landsat-5 dataset (from 1990 to 2010) to pretrain the model; the old sensor (from the 80s) has many sensor errors and noise; however, the huge amount of collected data over a large temporal span makes this sensor ideal to learn dynamics for many areas around the globe. Sentinel-2, which is more modern, has a higher revisit frequency and higher quality images (from 2015 until now), is used for testing and further fine-tuning in the following way: the Brazilian rivers are used for fine-tuning, to align the features learned from Landsat-5 to Sentinel-2 sensor, while Europe and USA are used for the testing. In this way, we collect around 1900 time series for testing and around 16000 for training, for a total of over 100 thousand single images.

\section{Tasks}
\begin{figure}[htb]
    \centering
    \begin{subfigure}[b]{0.3\linewidth}
        \centering
        \includegraphics[width=\linewidth]{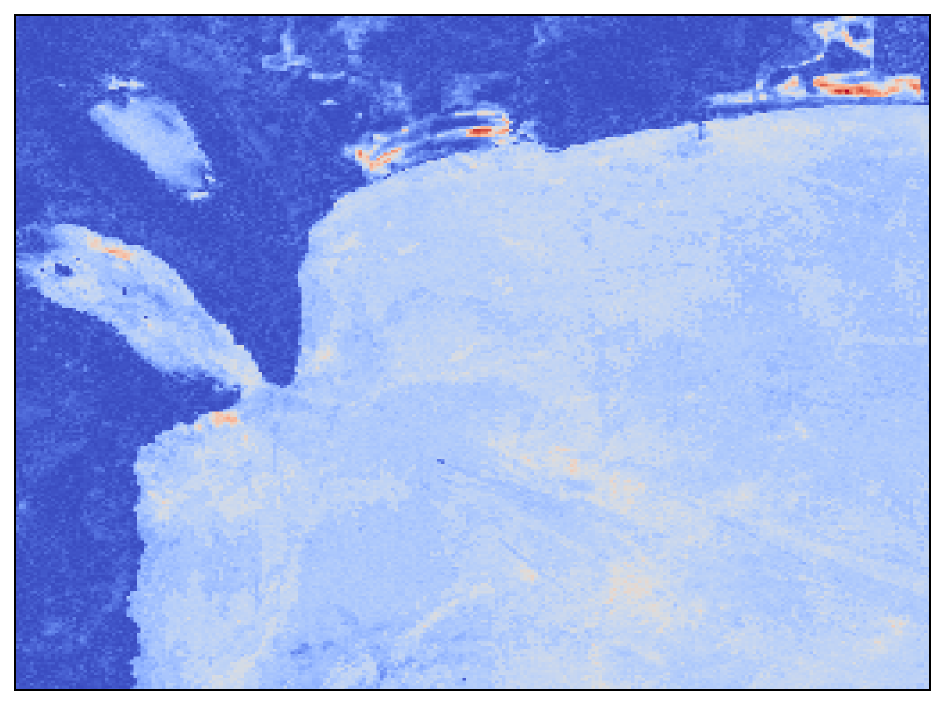} 
        \caption{Regression}
        \label{fig:regression_task}
    \end{subfigure}
    \begin{subfigure}[b]{0.3\linewidth}
        \centering
        \includegraphics[width=\linewidth]{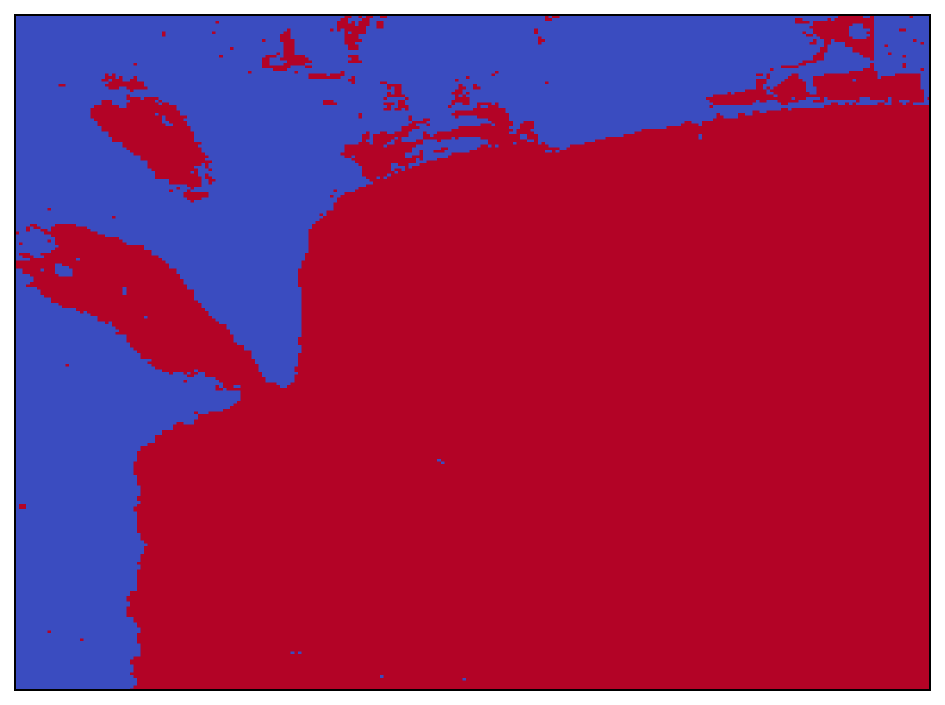} 
        \caption{Change}
        \label{fig:change_task}
    \end{subfigure}
    \begin{subfigure}[b]{0.3\linewidth}
        \centering
        \includegraphics[width=\linewidth]{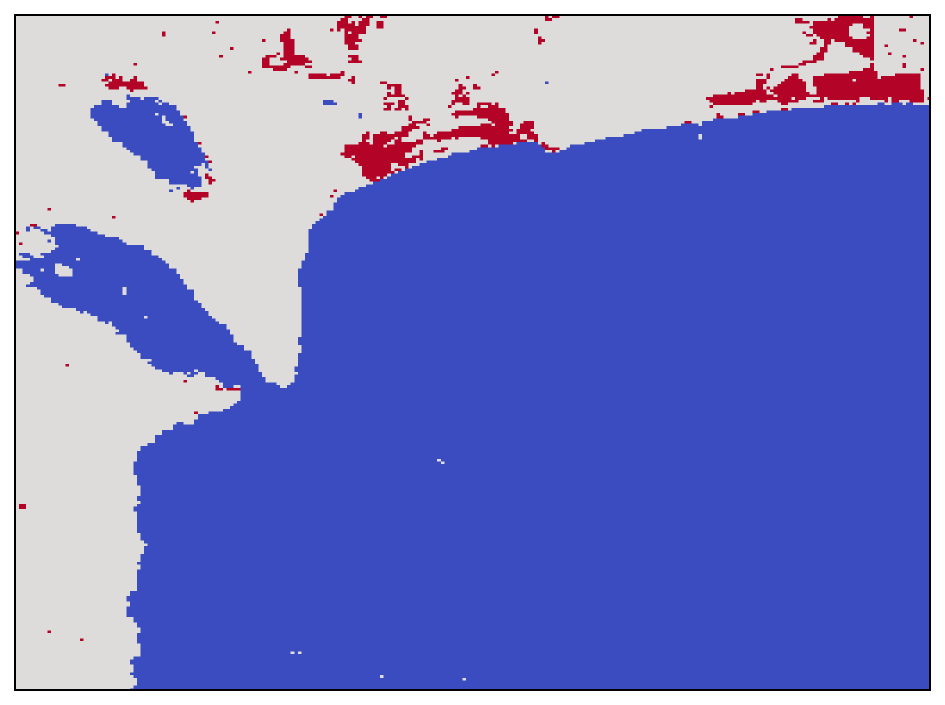} 
        \caption{Direction}
        \label{fig:direction_task}
    \end{subfigure}
    \caption{Visual example of tasks for Lake Tahoe. In regression, the values range from 0 to 2 (blue to red). In change detection, labels are no-change (blue) and change (red). In direction classification, labels are negative change (blue), no-change (grey), and positive change (red).}
    \label{fig:tasks_example}
\end{figure}

In this section, we delineate the target used in the proposed tasks and how each task is formulated. A visual example of the tasks is shown in \Cref{fig:tasks_example}.

\begin{figure}[htb]
    \centering
    \begin{tabular}{cc}
       \includegraphics[width=0.4\linewidth]{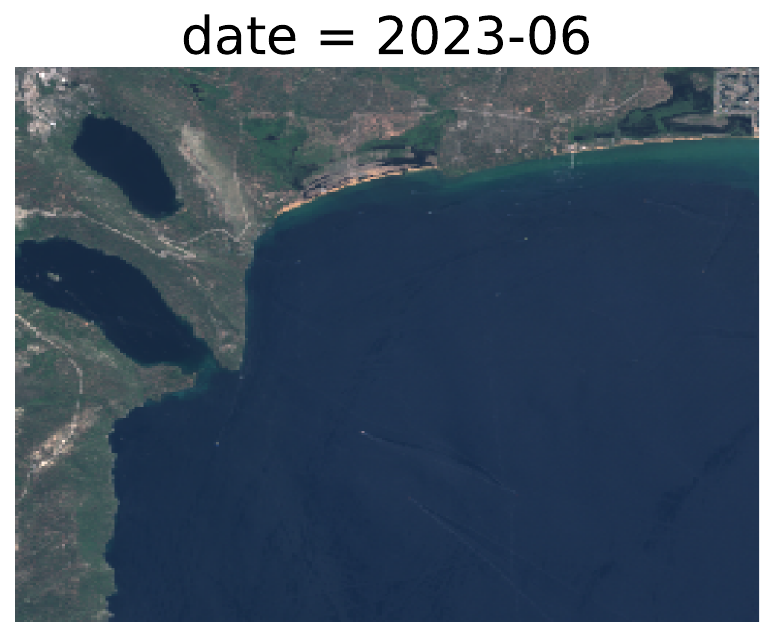}  &     \includegraphics[width=0.4\linewidth]{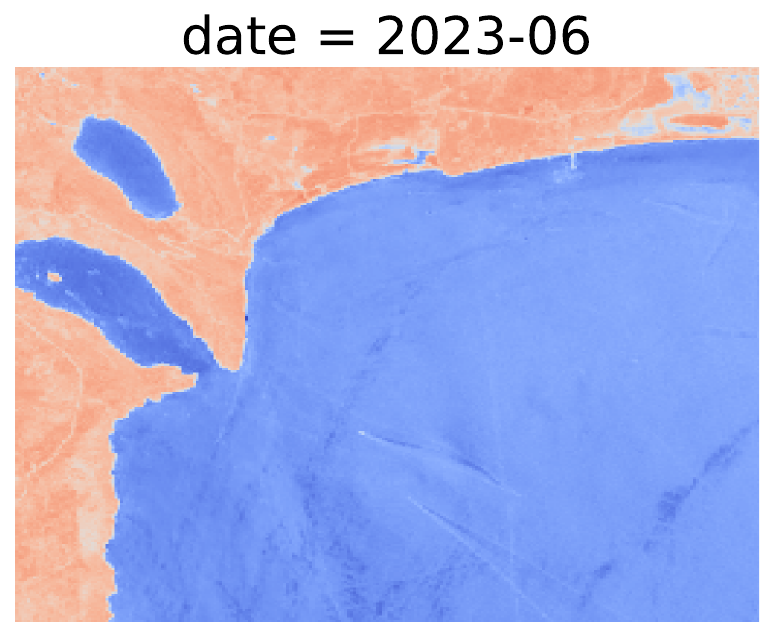} \\
       \includegraphics[width=0.4\linewidth]{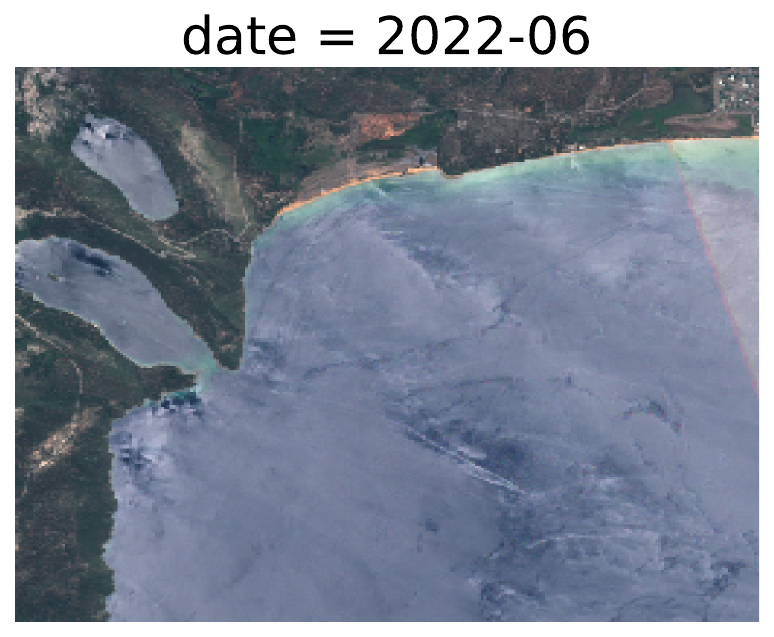}  &     \includegraphics[width=0.4\linewidth]{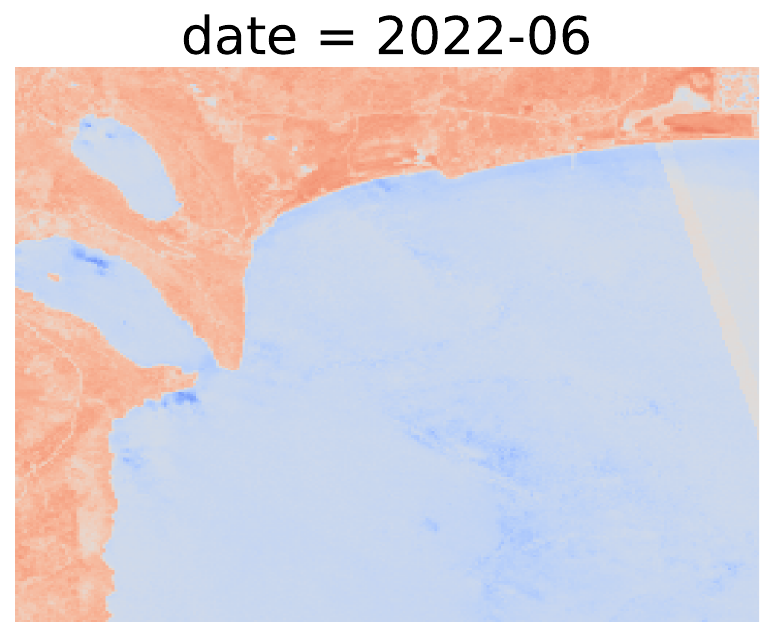} 
    \end{tabular}
    \caption{RGB sample at two different timesteps and the corresponding MNDWIs.}
    \label{fig:mndwi_sample}
\end{figure}

\subsection{General Target}
Given the difficulty in finding yearly annotations of water dynamics all over the world, we based our task on one renowned geo-index to detect water: MNDWI. A similar approach for NDVI was already explored \cite{Benson2024}. It is based on the physical properties of water, which is highly reflective in the green channel (G) and absorbs SWIR: $MNDWI = ({G - SWIR})/({G + SWIR})$.
This approach, while prone to noise, has the advantage of predicting directly the changes regarding physical properties (e.g., icing \cite{Warren2019}, turbidity \cite{Wojcik2022}, pollution \cite{Lu2019,Lu2020,zaugg2024}) rather than only focusing on water extents and avoiding a costly annotation process. An example can be seen in \Cref{fig:mndwi_sample}, where the icing process lowers the MNDWI values of the same area.

To avoid any inconsistency, due to the cloud, we apply cloud masking to invalid areas. Additionally, the imperfection of cloud masking and possible sensor errors is avoided with the following setting: given a past timeseries $P$ and a future timeseries $F$ of MNDWIs, our target $T$ is defined as: $T = median(P) - median(F)$, where the median is applied pixelwise over the time axis. In this way, instead of predicting the immediate, possibly noisy future, we target the future trend of the area. This pre-processing step is crucial for a large-scale analysis across diverse regions like the US, Europe, and Brazil, as it effectively reduces localized noise, accounts for minor short-term fluctuations in water levels or atmospheric interference, and smooths the MNDWI signal. Consequently, the subsequent change detection focuses on more persistent and significant alterations in water dynamics rather than ephemeral changes or sensor artifacts. Since MNDWI $\in \{-1, 1\}$, $T \in \{-2, 2\}$. The target distribution is strongly skewed (the median is 0.01, and the 75th percentile is 0.06) towards zero, as expected. %

\subsection{Change Detection}
The first proposed task is binary change detection, which can be framed as binary semantic segmentation. Given a timeseries $P$, a target $T$, and a threshold $t$ to define what we consider a relevant change, we create a binary mask $M_c = |T| > t$. The task focuses on creating a model to predict $M_c$.

\subsection{Direction Classification}
This task can be framed as a multiclass semantic segmentation task with 3 classes: negative, positive, or no change. Given a timeseries $P$, a target $T$, and a threshold $t$, we create a mask $M_d$ where a pixel $m_d$ is assigned to the negative change class if $m_d < t$, to the positive change class if $m_d > t$, otherwise it is assigned to the no change class. The task focuses on creating a model to predict $M_d$.

\subsection{Magnitude Regression}
The previous tasks assume the existence of a threshold $t$ to define relevant changes. However, it can be of interest to model every "small" change in the area. Given a timeseries $P$ and a target $T$, the task focuses on creating a model to regress the values of $|T|$. This task can be framed as pixel-wise regression. Preliminary experiments also tried to address the regression of $T$, but with little success, so we reported only these settings as baseline, leaving this last task for future work. 

\section{Methodology}
In this section, we first discuss our proposed baseline, and then we explain the regression loss we employed.

\subsection{AquaClimaTempo UNet}
\begin{figure}[htb]
    \centering
    \includegraphics[width=\linewidth]{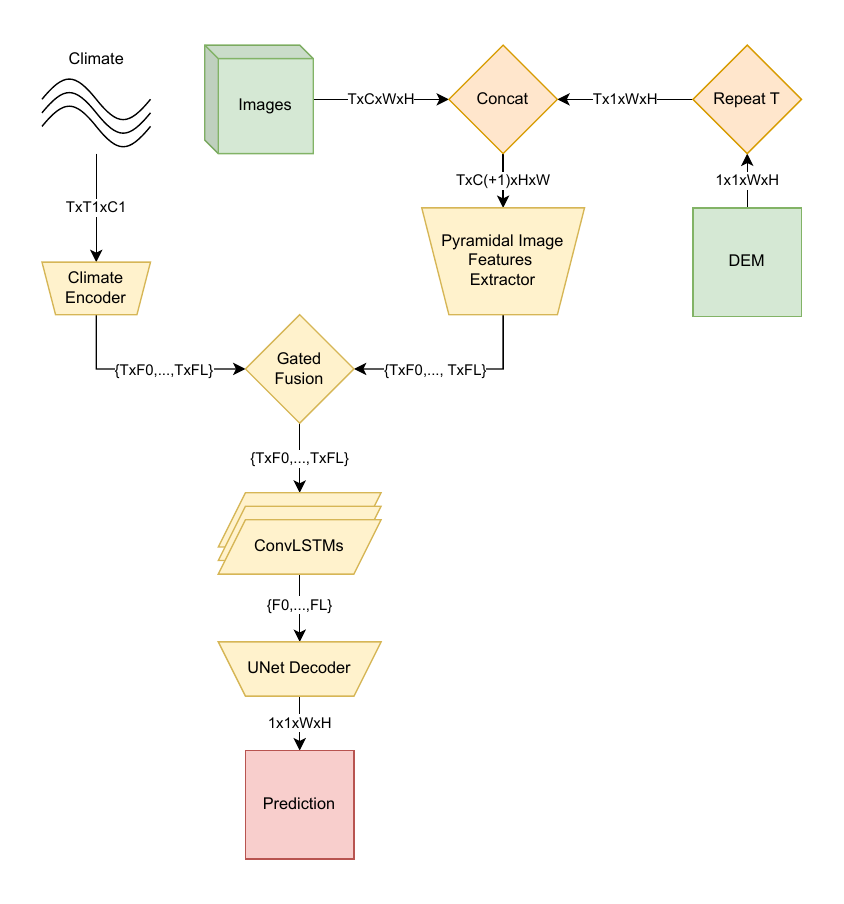}
    \caption{AquaClimaTempo UNet (ACTU) architecture. If DEM is provided, it is repeated once per sample in the image timeseries and concatenated along the channel axis. The \textit{Pyramidal Image Feature Extractor} provides multiscale embeddings. If a climate timeseries is provided, the \textit{climate encoder} provides multiscale embeddings which are \textit{gate fused} with the image embeddings. \textit{ConvLSTMs} provide multiscale embeddings for the timeseries, which are used in the \textit{UNet decoder} to provide the final prediction.}
    \label{fig:high-level-actu}
\end{figure}

The AquaClimaTempo UNet (ACTU) architecture is depicted in \Cref{fig:high-level-actu}. 
If DEM of shape $1 \times 1 \times W \times H$ is given, it is repeated $T$ times, one for each sample of the image timeseries $P$ to which is concatenated along the channel axis. The image time series of shape $T \times C \times W \times H$ (eventually $C + 1$, in case DEM is concatenated) is given to the \textit{Pyramid Image Feature Extractor} (e.g, ConvNext \cite{sanghyun2023}). It processes each image independently, and since it is pyramidal, it provides $L$ features, one for each level, of different shapes from $F_0$ to $F_L$.
If the climate variables are provided, they are one for each of the $T$ images and cover the past $T_1$ months of each image. The variables are $C_1$. The \textit{climate encoder} produces $L$ features with shapes from $F_0$ to $F_L$. The \textit{gated fusion} takes as input both the image and climate features at different levels and dynamically balances the contribution, outputting the same $L$ features of shapes from $F_0$ to $F_L$, one for each of the $T$ images. 
The \textit{ConvLSTM} layers flatten the time dimension, producing a representation for the timeseries for each of the $L$ levels of shape from $F_0$ to $F_L$.
Finally, the \textit{UNet decoder} takes the multilevel features and creates the final prediction mask by concatenating the features created by the expanding path with the ones obtained with the ConvLSTMs.

\subsubsection{Climate Encoder}
The climate encoder processes $X_{clim}$, one timeseries for each of the $T$ images, with length $T_1$ and $C_1$ features. In this way, we can incorporate historical trends with a finer timestep (i.e, monthly instead of yearly). They are independently processed and projected with a linear layer to create $F_{proj}$ (\Cref{eq:proj}). They are then processed to create $K_l$ representation to spatially match the image features using $L$ blocks composed of an initial Conv2D (with kernel 1) and GELU (\Cref{eq:first_up}), followed by $S$ series of nearest neighbor upsampling (with resize factor 2), Conv2D (with kernel 3), and GELUs (\Cref{eq:gen_up}).
\begin{align}
    F_{proj} = \text{Linear}(\Phi_{\text{LSTM}}(X_{clim})) \label{eq:proj} \\
    K_l^0 = \text{GELU}(\text{Conv2D}_{1\times1}(F_{proj})) \label{eq:first_up}\\
    K_l^s = \text{GELU}(\text{Conv2D}_{3\times3}(\text{Upsample}_{\times2}(K_l^{s - 1}))) \label{eq:gen_up}
\end{align}

\subsubsection{Gated Fusion}
The climate information can act differently based on the image itself, and also based on the area of the single image. To dynamically adapt the contribution of climate features over the image features, we apply a gate fusion. This solution provides a value in the range 0-1 for each element of the matrix to weight the contribution of optical and climate features dynamically. The weights are obtained by concatenating along the channels axis climate and optical features, and applying a Conv2D (with kernel 3), a ReLU, and Conv2D (with kernel 1), and finally a sigmoid to constrain the input in the range 0-1. Given the climate feature $K_l$ and the corresponding image feature $I_l$ at the level $L$, the gated fusion can be formulated:
\begin{align}
    Z = \text{Concat}(K_l, I_l) \\
    \alpha = \sigma(\text{Conv2D}_{1\times1}(\text{ReLU}(\text{Conv2D}_{3\times3}(Z)))) \\
    F_l = \alpha I_l + (1 - \alpha) K_l
\end{align}
Each $F_l$ is then used in the decoder the make the final prediction.

\subsection{Regression Loss}
Since the satellite resolution can vary and the problem shows a strong imbalance, L1 or L2 losses alone can be insufficient, as they are also sensitive to noise. Our regression loss makes use of HuberLoss as a starting point, but combines a multi-scale approach with a wavelet decomposition. 

\subsubsection{Multiscale Loss}
The multiscale loss $L_{MS}$, given a regression loss $L$, prediction $P$, and ground truth $T$ can be defined as:
\begin{equation}
    L_{MS}(P, T) = \frac{1}{M} \left(L(P, T) + \sum_{i=1}^{M} L(D_{s_i}(P), D_{s_i}(T)) \right)
\end{equation}
where $S = \{s_0, ..., s_M\}$ are $M$ different scale factors and $D_x$ is the downscale operation with factor $x$.

\subsubsection{Wavelet Loss}
The Discrete Wavelet Transform (DWT) decomposes both the prediction and the target into different frequency sub-bands, instead of directly comparing pixel values in the spatial domain. Compared to the Fourier transform, it is computationally more efficient ($O(N)$ compared to $O(N\log(N))$). This allows the loss to penalize errors at different scales and orientations (horizontal, vertical, diagonal). DWT decomposes the image with two coefficients: the approximation coefficients $Y_L$ (the low-frequency components, capturing its coarse structure) and detail coefficients $Y_H$ (the high-frequency components at $N$ levels and horizontal, vertical, diagonal orientations, capturing finer details). Given a regression loss $L$, prediction wavelet coefficients $Y_H^p$ and $Y_L^p$, and ground truth coefficients $Y_H^t$ and $Y_L^t$, the wavelet loss $L_W$ can be defined as:
\begin{align}
  L_L(Y_L^p, Y_L^t) = \text{mean}(L(Y_L^p, Y_L^t)) \\
  L_{H,i}(Y_{H,i}^p, Y_{H,i}^t) = \text{mean}(L(Y_{H,i}^p, Y_{H,i}^t)) \\
  L_W(Y_L^p, Y_L^t, Y_H^p, Y_H^t) = \alpha L_L(Y_L^p, Y_L^t) + \sum_{i=1}^{N} w_i \cdot L_{H,i}(Y_{H,i}^p, Y_{H,i}^t)
\end{align}
where $\alpha$ defines the weight of the low-frequency loss and $W = \{w_0, ..., w_N\}$ the weights of the high frequency losses.

The final loss is a weighted mean of the multiscale and wavelet losses: $L_T = \alpha L_{MS} + (1 - \alpha)L_W$.

\subsection{Explainable AI analysis}
The Explainable AI analysis investigates model behavior from two complementary perspectives.
First, we leverage climate-derived attributes to perform \textit{subgroup discovery and feature attribution}, aiming to reveal systematic performance disparities across interpretable climate conditions.
Second, we conduct a \textit{per-channel saliency analysis} to quantify the contribution of individual input modalities to the model’s predictions.

\subsubsection{Climate Subgroup Discovery and Feature Attribution}
Model performance in spatial machine learning can vary significantly across different environmental conditions. Aggregated metrics may conceal systematic failures concentrated in specific climatic regimes. To uncover and explain such disparities, we adopt a post hoc analysis framework based on subgroup discovery and feature attribution.

\paragraph{Climate Subgroup Discovery.}\label{par:subgroup} Let $\mathcal{D} = \{(x_i, y_i, \hat{y}_i)\}_{i=1}^{N}$ denote the evaluation dataset, where $x_i \in \mathcal{X}$ represents the input sample, $y_i \in \mathcal{Y}$ is the ground-truth label, and $\hat{y}_i \in \mathcal{Y}$ is the model prediction. Each $x_i$ is associated with a set of interpretable climate-derived attributes $A(x_i) = \{a_1, \ldots, a_k\}$ obtained via feature binning.

We define a \emph{subgroup} $S \subseteq \mathcal{D}$ as the set of samples satisfying a conjunction of attribute-value conditions:
\[
S = \{x_i \in \mathcal{D} \mid a_j(x_i) = v_j \quad \forall j \in J\},
\]
where $J \subseteq \{1, \ldots, k\}$ indexes selected attributes and $v_j$ denotes specific bin values.
To evaluate the behavior of a model over $S$, we use the notion of subgroup performance divergence~\cite{pastor2021divergent} defined as:
\[
\Delta_m(S) = m(S) - m(\mathcal{D}),
\]
where $m: \mathcal{D} \to \mathbb{R}$ is a scalar performance metric (e.g., precision, recall), $m(S)$ is the metric evaluated over the subgroup, and $m(\mathcal{D})$ is the global reference over the entire dataset.

We automatically identify the subgroups with large divergence scores using \texttt{DivExplorer}~\cite{pastor2021divergent}, which enumerates statistically significant subgroups under a minimum support constraint $\theta$. This process identifies climate conditions under which the model under- or over- performs.

\paragraph{Feature Attribution.} 
To explain which features contribute most to these performance deviations, we use the notion of Global Shapley values~\cite{pastor2021divergent}. The Global Shapley value is a generalization of the Shapley value~\cite{shapley:book1952} which estimates the contribution of each attribute-value to the divergence across all identified subgroups above the support constraint.
The higher the value, the more the attribute-value term contributes to the divergence in performance. A positive contribution of a term indicates that it is associated with a performance metric $m$ higher than the average on the overall dataset. We refer the reader to~\cite{pastor2021divergent} for its formal definition.

\subsubsection{Per-channel Saliency}

A central goal in XAI is to identify which input components most significantly influence a model's predictions. Beyond interpretability, this analysis has practical implications, such as reducing computational overhead by pruning less informative input channels. A common approach for estimating input relevance is perturbation-based and involves perturbing the input and measuring the resulting change in the model’s output. Perturbation-based techniques~\cite{zeiler2014visualizing, covertexplaining2021, breiman2001random} provide a straightforward method for attributing importance scores to input dimensions based on their effect on model behavior.
We adopt a perturbation-based strategy to compute \emph{per-channel saliency}, aimed at quantifying the relevance of each input channel. We adapted the method proposed in ~\cite{rege2025kan} to our time-series setting. Let \( x \in \mathbb{R}^{T \times C \times H \times W} \) denote the input tensor, where \( T \) is the number of temporal frames, \( C \) the number of channels, and \( H \times W \) the spatial resolution. For a given test sample, we first compute the model’s baseline prediction \( \hat{y} = \mathcal{M}(x, \text{DEM}, \text{Climate}) \), and evaluate it using a suite of performance metrics.
To assess the importance of channel \( c \in \{1, \ldots, C\} \), we generate a perturbed input \( x^{(-c)} \) by zeroing out the \( c \)-th channel across all time steps. We then recompute the model output as \( \hat{y}^{(-c)} = \mathcal{M}(x^{(-c)}, \text{DEM}, \text{Climate}) \). The saliency of channel \( c \) is defined as the change in a given performance metric \( m \):
$$
\Delta m_c = m(\hat{y}, y) - m(\hat{y}^{(-c)}, y),
$$
where \( y \) is the ground truth. A larger 
$\Delta m_c $ indicates that channel \( c \) has a greater influence on model performance, since its absence leads to a more substantial degradation in prediction quality.
We then average the per-channel saliency scores across the test set to obtain a dataset-level contribution.
This process is extended to the DEM input, which is ablated entirely to measure its global contribution.
Overall, this saliency analysis provides both local (per-sample) and global (dataset-level) interpretability,  helping to identify which input modalities most influence the model decisions in spatiotemporal tasks.

\section{Experimental Results}
In this section, we present the experimental settings and the results compared to simple statistical methods, namely constant prediction and persistence. Constant prediction is simply predicting that no change will happen in the future. Persistence for robustness is computed as the difference between the last known timestep and the median of the previous (thresholded with $t$ for classification).

\subsection{Experimental Settings}
ACTU is pretrained for 50 epochs on the Landsat subset, and fine-tuned for 20 epochs on Sentinel-2 with a cosine decaying learning rate scheduler with a 5\% warmup. The maximum learning rate is 5e-4 for the pretraining and 5e-6 for the fine-tuning. The batch size was set to 8. The loss for classification is a combo loss composed of generalized dice and focal losses. The vision backbone for the encoder is ConvNextV2 \cite{sanghyun2023}, in base (ACTU) and large (ACTU-L) versions. The length of the two image time series was set to 5. %
To avoid overwhelming the neural network with information (and deal with eventual multicollinearity and increasing costs), we select a subset of 5 climate variables (maximum temperature (tmmx), actual evapotranspiration (aet), runoff (ro), precipitation (pr), and soil moisture (soil)) that should be relevant for the task \cite{Habibi2025,Brocca2017,Ritter2011}. 
We also use these five variables for the analysis in~\Cref{par:subgroup}. Since this analysis requires categorical attributes to define subgroups, we discretized each variable into three interpretable bins. 
Specifically, we computed the standard deviation of each climate variable over the input time series to quantify temporal variability. These variability scores were discretized by frequency into three levels—\textit{low (L)}, \textit{medium (M)}, and \textit{high (H)}—thus enabling the construction of climate subgroups with distinct fluctuation profiles.
In our experiments, we threshold at $t=0.1$ (i.e., $\sim$85th percentile) to remove possible sensor noises, atmospheric effects, phenological changes, and coregistration errors. A key requirement for training a neural network is the generation of a consistent change mask across the entire diverse study area. A fixed threshold ensures that the definition of `change' is uniform. While this approach is necessary for large-scale change detection, it is acknowledged that the precise magnitude of MNDWI difference that corresponds to a `relevant' real-world change can still exhibit some variability due to the diverse nature of water bodies, atmospheric conditions, and local landscape characteristics across the US, Europe, and Brazil.

\subsection{Evaluation Metrics}
We evaluate classification tasks using precision (P), recall (R), and F1-score (F) for each class. Since the regression problem is pixel-wise, and many areas have values near zero, it can be considered ``unbalanced''. We evaluate the regression quality with Mean Absolute Error (MAE) and the Pearson Correlation (PC). We also compute MAE on the top-10 (MAE@10) and top-20 (MAE@20) highest valued pixels to account for the imbalance of the values. We threshold the regressed values at $t=0.1$ (as for classification) and $t=0.2$, where we compute precision (P@t), recall (R@t), and F1-score (F@t). This allows us to assess the trade-offs between detecting relevant pixels and the accuracy of those detections at different sensitivity levels.

\subsection{Change Detection}
\Cref{tab:binary_results} presents the binary change detection results, highlighting the performance of various model configurations. All ACTU model variants demonstrate a substantial and statistically significant improvement in detecting changes (CHG) compared to the \textit{Constant} and \textit{Persistence} baselines. For instance, the baselines achieve F1-scores of 0 and 34.98, respectively, whereas all ACTU configurations surpass an F1-score of 45 for this class. The inclusion of climate variables (C) with the base ACTU model improves the no-change class (NoCHG) F1-score but results in a slight decrease in the CHG F1-score. Conversely, incorporating only DEM data (D) enhances the CHG F1-score, the highest among the standard ACTU variants for this metric, while also slightly improving NoCHG performance. When both DEM and climate data are utilized, the model achieves the highest CHG Recall (62.33), proving its superior capability in identifying actual change instances, though its F1-score (48.67) is marginally lower than the DEM-only configuration. The larger backbone model, ACTU-L, shows a modest improvement in CHG precision. Generally, a larger backbone does not seem to provide consistent improvements over the base version.

\begin{table}[htb]
    \centering
    \caption{Change detection results for models optionally using DEM (D) and climate variables (C). * indicates statistically significant difference (p < 0.01) with respect to persistence according to the t-test. ° indicates the statistical difference comparing ACTU-L with the same configuration of ACTU.}
    \label{tab:binary_results}
    \resizebox{\linewidth}{!}{\begin{tabular}{@{}l|cc|ccc|ccc@{}}
\toprule
            &   &   & \multicolumn{3}{c|}{No Change (NoCHG)} & \multicolumn{3}{c}{Change (CHG)} \\ \midrule
Model       & D & C & P        & R       & F      & P       & R       & F       \\ \midrule
Constant    & N & N & 81.54    & \textbf{100}     & \textbf{89.25}   & 0       & 0       & 0       \\
Persistence & N & N & 88.73    & 41.64   & 54.07   & 23.86   & \textbf{81.77}   & 34.98   \\ \midrule
ACTU        & N & N & 90.51*   & 82.78*  & 85.66*  & 44.87*  & 60.65*  & 48.79*  \\
ACTU        & N & Y & 88.92*   & 85.86*  & 86.6*   & 45.45*  & 53.1*   & 45.83*  \\
ACTU        & Y & N & \textbf{90.57}    & 82.89*  & 85.75*  & 45.19*  & 61.01*  & \textbf{49.38*}  \\
ACTU        & Y & Y & 90.53*   & 81.71*  & 85.08*  & 43.68*  & 62.33*  & 48.67*  \\ \midrule
ACTU-L      & N & N & 90.03°   & 84.3°   & 86.43°  & \textbf{45.46°}  & 57.34°  & 47.94°  \\
ACTU-L      & Y & Y & 89.2°    & 84.95°  & 86.37°  & 45.03°  & 54.39°  & 46.49°  \\ \bottomrule
\end{tabular}
}
\end{table}

\subsection{Direction Classification}
\begin{table*}[htb]
    \centering
    \caption{Direction classification results for models optionally using DEM (D) and climate variables (C). * indicates statistically significant difference (p < 0.05) with respect to persistence according to the t-test. ° indicates the statistical difference comparing ACTU-L with the same configuration of ACTU.}
    \label{tab:direction_results}
    \begin{tabular}{@{}l|cc|ccc|ccc|ccc@{}}
\toprule
            &   &   & \multicolumn{3}{c|}{Negative Change (NEG)} & \multicolumn{3}{c|}{No Change (NONE)} & \multicolumn{3}{c}{Positive Change (POS)} \\ \midrule
Model       & D & C & P        & R       & F       & P        & R       & F       & P       & R       & F       \\ \midrule
Constant    & N & N & 0        & 0       & 0       & 81.54    & \textbf{100}     & \textbf{89.25}   & 0       & 0       & 0       \\
Persistence & N & N & 14.94    & 47.37   & 17.48   & 88.73    & 41.64   & 54.07   & 9.25    & \textbf{31.18}   & 11.61   \\ \midrule
ACTU        & N & N & 27.5*    & \textbf{49.38*}  & \textbf{30.27*}  & \textbf{90.23*}   & 84.58*  & 86.67*  & 27.73*  & 19.84*  & 19.47*  \\
ACTU        & N & Y & \textbf{29.84*}   & 36.53*  & 27.75*  & 88.44    & 87.5*   & 87.42*  & 26.16*  & 24.12*  & \textbf{21.38*}  \\
ACTU        & Y & N & 28.18*   & 34.43*  & 25.81*  & 87.73*   & 88.65*  & 87.54*  & 27.39*  & 20.25*  & 19.09*  \\
ACTU        & Y & Y & 28.36*   & 37.35*  & 27.28*  & 88.18*   & 88.17*  & 87.6*   & 27.98*  & 22.06*  & 20.77*  \\ \midrule
ACTU-L      & N & N & 28.42°   & 38.21°  & 27.62°  & 88.5°    & 88.01°  & 87.76°  & \textbf{28.25°}  & 22.04°  & 20.88°  \\
ACTU-L      & Y & Y & 27.52°   & 34.84°  & 25.31°  & 88.15    & 88.09   & 87.55   & 27.13°  & 21.28°  & 19.44°  \\ \bottomrule
\end{tabular}

\end{table*}

\Cref{tab:direction_results} details the performance for the direction classification task, categorizing changes into negative (NEG), no change (NONE), or positive (POS). All ACTU model variants achieve statistically significant and considerable improvements over the \textit{Constant} and \textit{Persistence} baselines. This is particularly evident for POS and NEG, where the \textit{Persistence} achieves 17.48 and 11.61, respectively, compared to ACTU 30.27 and 19.47. All models excel at identifying NONE, with ACTU variants consistently reaching F1-scores around 86-87. However, accurately classifying the direction of change (POS and NEG) is inherently more challenging, as reflected by their lower F1-scores compared to NONE across all models.
Introducing climate variables notably improves the F1-score for POS, though it slightly reduces performance for NEG. Conversely, adding DEM data alone does not yield a clear F1-score improvement for either change direction class, slightly decreasing NEG and POS F1-scores. The combined use of DEM and climate data results in a robust NONE F1-score (87.6) and a POS F1-score (20.77). The larger ACTU-L model shows modest F1 improvements for NONE (87.76) and POS (20.88) over its smaller counterpart, but a decrease for NEG. These results suggest that while ACTU is strongest for detecting negative changes, incorporating climate data is particularly beneficial for identifying positive changes. The overall task of precise direction classification remains complex, with input data types showing varied impacts on different change categories.

\subsection{Magnitude Regression}
\Cref{tab:regression_results} details the magnitude regression performance, where all ACTU model variants demonstrate statistically significant and substantial improvements over the \textit{Constant} and \textit{Persistence} baselines across all reported metrics. The standard ACTU model achieves a low MAE of 0.0261 and a PC of 46.45. While the incorporation of DEM (D), climate (C) variables, or both tends to slightly increase overall MAE and decrease PC, the inclusion of DEM markedly improves the F1-scores when regression outputs are thresholded to identify significant changes. Specifically, F@0.1 and F@0.2 improve due to enhanced recall. This indicates that while the base model excels at general magnitude prediction, DEM input is particularly beneficial for more accurately identifying pixels undergoing substantial change. The larger backbone model, ACTU-L, emerges as the top-performing configuration. It improves MAE for the top 10\% and 20\% highest magnitude changes and achieves the highest F1-scores for thresholded significant changes. Although its overall MAE is marginally higher than ACTU, its PC is slightly better. In contrast, ACTU-L with climate and dem, while improving general MAE and PC over its smaller counterpart ACTU, does not reach the thresholded performance levels of ACTU-L.

\begin{table}[htb]
    \centering
    \caption{Magnitude regression results for models optionally using DEM (D) and climate variables (C). * indicates statistically significant difference (p < 0.05) with respect to persistence according to the t-test. ° indicates the statistical difference comparing ACTU-L with the same configuration of ACTU.}
    \label{tab:regression_results}
    \begin{subtable}{\linewidth}
        \centering
        \begin{tabular}{@{}l|cc|cccc@{}}
\toprule
Model       & D & C & MAE    & MAE@10 & MAE@20 & PC      \\ \midrule
Constant    & N & N & .0351  & .142   & .1038  & -       \\
Persistence & N & N & .1281  & .1892  & .171   & 31.81    \\ \midrule
ACTU        & N & N & \textbf{.0261*} & .0873* & .0611* & 46.45*  \\
ACTU        & N & Y & .0266* & .0911* & .0639* & 44.4* \\
ACTU        & Y & N & .0297* & .0886* & .0643* & 44.46*   \\
ACTU        & Y & Y & .0315* & .088*  & .0634* & 41.41*   \\ \midrule
ACTU-L      & N & N & .0275° & \textbf{.0843°} & \textbf{.0589°} & \textbf{46.62}  \\
ACTU-L      & Y & Y & .0282° & .0923° & .066°  & 43.45° \\ \bottomrule
\end{tabular}

    \end{subtable}

    \vspace{1em}
    
    \begin{subtable}{\linewidth}
        \resizebox{\linewidth}{!}{\begin{tabular}{@{}l|cc|cccccc@{}}
\toprule
Model       & D & C & P@0.1  & R@0.1  & F@0.1  & P@0.2  & R@0.2  & F@0.2  \\ \midrule
Constant    & N & N & 0      & 0      & 0      & 0      & 0      & 0      \\
Persistence & N & N & 23.86  & \textbf{81.77}  & 34.98  & 13.38  & \textbf{75.72}  & 20.25  \\ \midrule
ACTU        & N & N & \textbf{51.97*} & 41.61* & 43.04* & 45.27* & 24.48* & 26.85* \\
ACTU        & N & Y & 51.32* & 41.59* & 42.77* & 42.34* & 18.12* & 21.02* \\
ACTU        & Y & N & 49.36* & 45.09* & 43.64* & 40.67* & 28.26* & 27.8*  \\
ACTU        & Y & Y & 46.92* & 45.18* & 42.67* & 39.57* & 27.91* & 27.3*  \\ \midrule
ACTU-L      & N & N & 50°    & \textbf{47.19°} & \textbf{45.61°} & 44.45° & 27.55° & \textbf{28.65°} \\
ACTU-L      & Y & Y & 51.63° & 38.48° & 40.65° & 43.24° & 21.41° & 23.41° \\ \bottomrule
\end{tabular}
}
    \end{subtable}
\end{table}

\subsection{Ablation Studies on regression loss}
To understand the contribution of the proposed composed loss $L_T$, we report in \Cref{tab:regression_loss_ablation} the comparison with its loss components alone and the employment of a standard regression loss (Huber loss, the same used in the other derivative losses). We compare the ACTU without any additional information for simplicity.
Comparing the Huber loss ($L$) to the multiscale version ($L_{MS}$), $L_{MS}$ provides better recalls and so better F1-scores, and lower values of MAEs in the top-highest values.
The wavelet loss ($L_W$) provides good performance in regression, but struggles with classification metrics. This is probably due to the frequency domain, which lacks any spatial information.
The linear combination $L_T$ proves its benefits in regression metrics when looking at MAE and PC (at least +2\% improvement). Looking at classification metrics, it enhances the precision while affecting the recall. The F@0.1 remains not significantly affected by the recall loss, while F@0.2 is enhanced. It can be easily seen that it blends the highest recall of spatial loss ($L_{MS}$), with the highest precision of frequency loss ($L_W$), providing a balanced contribution of both. This loss proved to be a good alternative; still, a more extensive search could be performed in the future to select even better hyperparameters.

\begin{table}[htb]
    \caption{Comparison between the wavelet loss ($L_W$), multiscale loss ($L_{MS}$), their combination $L_T$, and standard application of the regression loss ($L$). * indicates statistically significant difference (p < 0.01) with respect to $L_T$ according to the t-test.}
    \label{tab:regression_loss_ablation}
    \begin{subtable}{\linewidth}
        \centering
        \begin{tabular}{@{}l|cccc@{}}
\toprule
Model & MAE            & MAE@10          & MAE@20          & PC                       \\ \midrule
$L_T$ & \textbf{.0261} & .0873           & .0611           & \textbf{46.45}           \\ \midrule
$L$ & .029*          & .0861*          & .06*            & 42.9*                             \\
$L_{MS}$    & .0291*         & \textbf{.0839*} & \textbf{.0585*} & 43.82*                  \\
$L_W$    & .0285*         & .1047*          & .0765*          & 42.1*                  \\ \bottomrule
\end{tabular}

    \end{subtable}
    
    \vspace{1em}
    
    \begin{subtable}{\linewidth}
        \centering
        \begin{tabular}{@{}l|cccccc@{}}
\toprule
Model & P@0.1          & R@0.1           & F@0.1          & P@0.2          & R@0.2         & F@0.2          \\ \midrule
$L_T$ & \textbf{51.97} & 41.61           & 43.04          & \textbf{45.27} & 24.48         & \textbf{26.85} \\ \midrule
$L$ & 46.14*         & 45.65*          & 42.25          & 39.89*         & 24.25         & 24.61*         \\
$L_{MS}$ & 45.85*         & \textbf{49.84*} & \textbf{44.83} & 39.38*         & \textbf{27.2} & 26.75*         \\
$L_W$    & 51.24          & 25.11*          & 29.88*         & 38.51*         & 16.24*        & 18.42*         \\ \bottomrule
\end{tabular}

    \end{subtable}
\end{table}

\section{Analysis and Discussion}
In this section, we present the insights derived from the XAI analysis.
We begin with the Climate Subgroup Discovery and Feature Attribution, which shows how the model’s performance changes according to climate variations and highlights its reliance on specific climate variables.
We then examine the Per-channel Saliency to assess the relative importance of the individual spectral bands and DEM.

\begin{table}[h]
    \centering
    \caption{
Lakes associated with the worst-performing climate subgroups across the three tasks: change detection (C), direction classification (D), and regression (R), with the affected classes (for C and D) and MAE metric for R.
}
    \label{tab:worst_lakes}
    \begin{tabular}{lccc}
    \toprule
    Lake & (C) & (D) & (R)   \\
      \midrule
    Great Salt Lake         &  CHG        &  POS        & MAE  \\
    Utah Lake               &  CHG        &  POS        & MAE \\
    Rainy Lake              &  NoCHG        &  NONE            & \textemdash  \\
    Woods Lake              &  NoCHG        &  NONE            & \textemdash  \\
    Lake Texoma             & \textemdash             &  POS        & MAE \\
    Pyramid Lake            & \textemdash             &  POS        & MAE \\
    \bottomrule
    \end{tabular}
\end{table}

\subsection{Climate Subgroups and Feature Attribution}
\textit{Climate Subgroups.}
We identify climate subgroups that consistently challenge the model across different tasks. These understandings enable us to outline critical samples characterized by hard-to-learn environmental patterns. 
We first perform the subgroup discovery over the five climate variables, discretized according to their intra-series variability.
For each task and class, we extract the subgroup exhibiting the highest divergence, as detailed in Appendix ~\ref{appendix_subgroups}. 
We then conduct a comparative analysis of these subgroups by inspecting the samples they encompass, in order to uncover recurring problematic regions across tasks. 
We show the findings in \Cref{tab:worst_lakes}.
Notably, there is strong agreement across all the tasks in the difficulty of predicting that the area will change in the regions of Great Salt Lake and Utah Lake, as they appear in the worst-performing subgroups for both change, direction, and regression models. 
Conversely, Rainy Lake and Woods Lake are particularly problematic when the model attempts to predict stable conditions (i.e., no change), suggesting that temporal climate fluctuations in these areas might mimic weak change signals. Additionally, Lake Texoma and Pyramid Lake consistently appear in subgroups associated with poor performance in both the regression task and the detection of positive changes, pointing to a possible shared climate signal that the model struggles to generalize across these scenarios.

\paragraph{Feature Attribution.}
We compute the Global Shapley values to quantify the contribution of each attribute-value pair to the overall divergence. This allows us to identify which factors are most responsible for performance variations across subgroups. 
\Cref{tab:climate_drivers} summarizes the two \textit{Most} and \textit{Least} contributing levels of climate variability for each of the three tasks. In the change detection task, precipitation (pr) and soil moisture (soil), particularly under conditions of low or high variability, consistently emerge as strong contributors to model performance.
Maximum temperature (tmmx) also appears repeatedly across all tasks, with its variability positively correlated with higher prediction accuracy. In contrast, high variability in evapotranspiration (aet) is frequently among the least contributing factors, indicating that it may introduce instability or ambiguity that the model struggles to effectively capture.
These findings indicate that model performance is not uniformly distributed across climatic regimes and that temporal variability in certain features (e.g., soil, aet) can lead to systematic failure modes. 
Identifying failure-prone climate subgroups not only enhances our understanding of climate feature relevance 
but also provides actionable insights for improving model robustness through targeted data augmentation, domain adaptation, and climate-aware validation strategies.

\subsection{Per-channel Saliency}
\begin{figure}[htb]
    \centering
    \includegraphics[width=\linewidth]{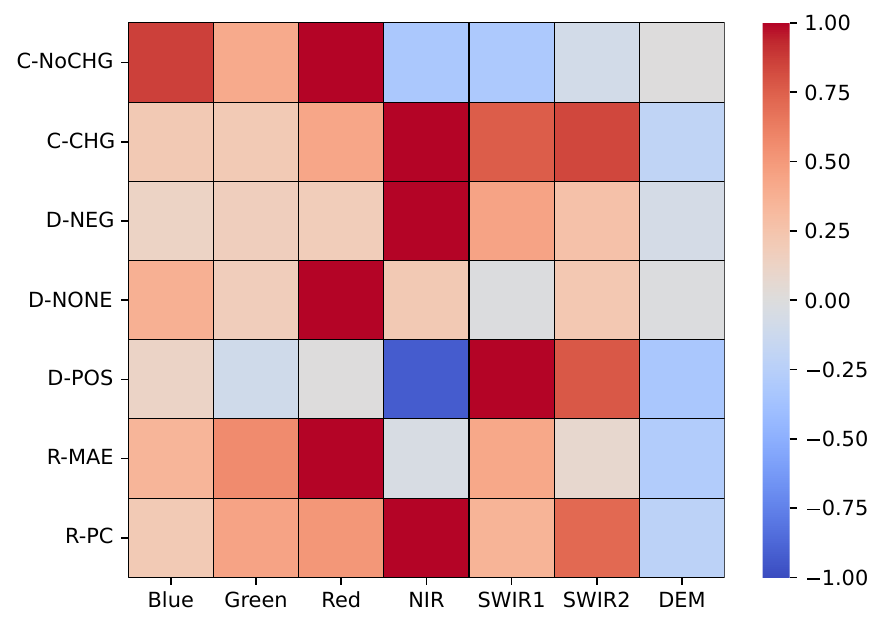}
    \caption{Row-Normalized Per-Channel Saliency related to change detection (C) and direction classification (D), F1-Score for each class, and to MAE and Pearson Correlation (PC) of regression (R).}
    \label{fig:saliency}
\end{figure}

We perform a per-channel saliency analysis to evaluate the contribution of each input modality.
For change detection (C) and direction classification (D), we computed the average drop in F1 score resulting from the ablation of individual channels. 
For the regression task (R), we instead measured the change in Mean Absolute Error (MAE) and Pearson Correlation (PC). 
We summarize the saliency scores in \Cref{fig:saliency}. %
All the saliency scores are row-normalized between -1 (confusing channel) and 1 (important channel) to allow for consistent visual comparison, where 0 indicates an irrelevant channel. For the MAE, we normalized its negative value to have a consistent interpretation. 
The models evaluated included the DEM as an additional input.

In the change detection task, NIR and SWIR channels stand out as the most informative for detecting change events (C-CHG, second row), while RGB channels play a greater role in predicting areas with no change (C-NoCHG, first row). This pattern largely holds in the direction classification task as well. Interestingly, although the NIR channel supports detection of negative changes (D-NEG), it appears to hinder the identification of positive changes (D-POS), revealing a class-dependent interaction with this spectral band.
In the regression task, performance improves when the model has access to the full spectrum of channels. %
For R-PC, NIR and SWIR maintain their prominence, yet the RGB channels also contribute consistently, suggesting a beneficial multispectral synergy.

This analysis highlights the distinct and complementary roles of spectral bands across tasks. NIR and SWIR are crucial for detecting dynamic changes and maintaining high correlation with ground truth signals, while RGB channels remain essential for stable predictions and class discrimination. Moreover, the varying impact of NIR on different direction classes underscores the importance of task-specific channel sensitivity when designing interpretable Earth observation models.

\begin{table}[htb]
    \centering
    \caption{Feature Attribution through Global Shapley values. For each of the tasks, we report the two best and worst contributing climate values (H=high, M=medium, L=low).}
    \label{tab:climate_drivers}
    \begin{tabular}{@{}l|l|ll|ll@{}}
    \toprule
                       &         & \multicolumn{2}{c|}{Most Contributing} & \multicolumn{2}{c}{Least Contributing} \\ \midrule
    \multirow{2}{*}{C} & NoCHG & pr=H        & soil=H      & ro=H        & aet=H       \\
                       & CHG & ro=M       & soil=L      & soil=H      & aet=M      \\ \midrule
    \multirow{3}{*}{D} & NEG & pr=L        & tmmx=L      & pr=M       & pr=H        \\
                       & NONE & pr=H        & soil=H      & ro=H        & aet=H       \\
                       & POS & tmmx=L      & aet=L       & ro=H        & aet=H       \\ \midrule
    \multirow{2}{*}{R} & MAE     & soil=H      & tmmx=H      & ro=H        & aet=H       \\
                       & PC      & tmmx=L      & tmmx=H      & aet=M      & soil=H      \\ \bottomrule
    \end{tabular}
\end{table}

\section{Conclusion}
\textsc{HydroChronos} and our findings lay a foundation for advancing predictive spatio-temporal modeling in hydrology, fostering innovative approaches to proactive water resource management, and ultimately contributing to more sustainable and resilient water futures in an era of significant environmental change. Our dataset provides the first large-scale effort to map water evolution over the years, and ACTU provides a baseline that integrates not only visual features but also climate variables. 
Explainable AI analysis identified salient spectral bands like NIR and SWIR for change detection and varied impacts of climate variables. Additionally, our analysis provides experts with insight into the flaws and strengths of our model, guiding future works. In future works, additional tasks (e.g., MNDWI regression) should be investigated, and a fine-grained analysis of the relations between climate and visual features should be performed.

\bibliographystyle{ACM-Reference-Format}
\bibliography{bibliography}

\appendix

\section{Additional Training Settings}
The timeseries of images could have a variable length because of missing images for a given timestep of a minimum quality. We kept only the time series with at least 80\% of the timesteps. The input timeseries is zero-imputed to ensure a regular timestep (zero is assigned to no-data on the original image sources). The target is calculated between the two non-imputed time series to avoid introducing additional noise.  Since the satellites have different resolutions and quality, we apply random rotation, random translation, and random resize crop to ensure robustness, generalizability, and possible minor misalignment. All input data is standardized. We normalize the target to range -1 to 1. $\alpha = 0.5$ for the loss $L_T$. 

\section{Climate Subgroups } \label{appendix_subgroups}

Table~\ref{tab:appendix_subgroups} reports the most divergent climate subgroup identified for each task and class. These subgroups correspond to combinations of climate variability levels—discretized into \textit{low (L)}, \textit{medium (M)}, and \textit{high (H)}—across five key variables: precipitation (pr), temperature (tmmx), runoff (ro), evapotranspiration (aet), and soil moisture (soil).

Each subgroup reflects a set of environmental conditions under which the model exhibits the greatest divergence in performance, highlighting systematic weaknesses that recur across samples.

Alongside each subgroup, we report the corresponding Divergence score, which quantifies the severity of the model’s underperformance on that specific subgroup. 
For example, the first group has a F1-score for the NEG class by 0.24 lower than the average.
All the divergence scores of the reported subgroups are statistically significant as computed via the Welch-t test as defined in~\cite{pastor2021divergent}.

\begin{table}[ht]
\centering
\small
\caption{Most divergent climate subgroups for each task and class with associated divergence scores indicating the severity of model underperformance.}

\label{tab:appendix_subgroups}
\begin{tabular}{ll>{\raggedright\arraybackslash}p{3.5cm}r}
\toprule
\textbf{Task} & \textbf{Class} & \textbf{Most Divergent Subgroup} & \textbf{Divergence $\Delta$} \\
\midrule
D & NEG & \{pr\_bin=H, tmmx\_bin=H, ro\_bin=L\} & -0.24 \\
D & NONE & \makecell[tl]{\{aet\_bin=H, tmmx\_bin=H, soil\_bin=L,\\ pr\_bin=M, ro\_bin=H\}} & -0.18 \\
D & POS & \makecell[tl]{\{soil\_bin=M, pr\_bin=M,\\ aet\_bin=L, tmmx\_bin=M\}} & -0.14 \\
C & NoCHG & \makecell[tl]{\{soil\_bin=L, pr\_bin=M, ro\_bin=H,\\ aet\_bin=H, tmmx\_bin=H\}} & -0.18 \\
C & CHG & \makecell[tl]{\{tmmx\_bin=M, ro\_bin=L, pr\_bin=L,\\ soil\_bin=M, aet\_bin=M\}} & -0.22 \\
R & MAE & \makecell[tl]{\{tmmx\_bin=M, aet\_bin=L, pr\_bin=M,\\ soil\_bin=M, ro\_bin=L\}} & 0.016 \\
\bottomrule
\end{tabular}
\end{table}

\end{document}